\documentclass[conference]{IEEEtran}
\IEEEoverridecommandlockouts
\usepackage{cite}
\usepackage{amsmath,amssymb,amsfonts}
\usepackage{algorithmic}
\usepackage{graphicx}
\usepackage{textcomp}
\usepackage{xcolor}
\usepackage[utf8]{inputenc}
\usepackage[T1]{fontenc}
\usepackage[hidelinks]{hyperref}
\def\BibTeX{{\rm B\kern-.05em{\sc i\kern-.025em b}\kern-.08em
    T\kern-.1667em\lower.7ex\hbox{E}\kern-.125emX}}

\usepackage[many]{tcolorbox}
\tcbset{myprompt/.style={subtitle style={colback={blue3!20}}}}

\definecolor{blue1}{rgb}{0,0.13,0.25}
\definecolor{blue2}{rgb}{0,0.22,0.42}
\definecolor{blue3}{rgb}{0,0.33,0.61}
\definecolor{blue4}{rgb}{0,0.4,0.75}
\definecolor{blue5}{rgb}{0,0.47,0.87}
\definecolor{blue6}{rgb}{0.11,0.56,0.95}

\newtcolorbox{taskbox}[2][]{%
    enhanced,breakable,
    colframe=blue3!40,
    colback=blue5!5,
    arc=1mm,
    outer arc=1mm,
    fontupper=\small,
    fontlower=\small,
    coltitle=blue1,
    fonttitle=\bfseries,    
    boxsep=1mm,
    left=0mm,
    right=0mm,
    top=0mm,
    bottom=0mm,
    before={\noindent},
    segmentation style={solid, blue3},
    title=#2,%
    #1
}

\begin{document}

\title{From Big to Small Without Losing It All:\\Text Augmentation with ChatGPT\\for Efficient Sentiment Analysis
}

\author{\IEEEauthorblockN{1\textsuperscript{st} Stanisław Woźniak}
\IEEEauthorblockA{\textit{Department of Artificial Intelligence} \\
\textit{Wrocław University of Science and Technology}\\
Wrocław, Poland \\
stanislaw.wozniak@pwr.edu.pl}
\and
\IEEEauthorblockN{2\textsuperscript{nd} Jan Kocoń}
\IEEEauthorblockA{\textit{Department of Artificial Intelligence} \\
\textit{Wrocław University of Science and Technology}\\
Wrocław, Poland \\
jan.kocon@pwr.edu.pl}

}

\maketitle

\begin{abstract}
In the era of artificial intelligence, data is gold but costly to annotate. The paper demonstrates a groundbreaking solution to this dilemma using ChatGPT for text augmentation in sentiment analysis. We leverage ChatGPT's generative capabilities to create synthetic training data that significantly improves the performance of smaller models, making them competitive with, or even outperforming, their larger counterparts. This innovation enables models to be both efficient and effective, thereby reducing computational cost, inference time, and memory usage without compromising on quality. Our work marks a key advancement in the cost-effective development and deployment of robust sentiment analysis models.
\end{abstract}



\section{Introduction}

The burgeoning field of artificial intelligence (AI) is fueled by data—vast, often labeled datasets that serve as the training ground for machine learning models. Yet, collecting and annotating this data can be prohibitively expensive and time-consuming, creating a bottleneck in developing and deploying AI technologies. Parallel to these challenges, large language models like ChatGPT have attracted considerable attention for their remarkable generative capabilities, simulating human-like conversation and problem-solving skills \cite{mao2023gpteval,amin38will,kocon2023chatgpt}.

In light of these developments, we explore a synergistic approach that leverages the generative prowess of ChatGPT to enhance the efficiency and performance of smaller machine-learning models in sentiment analysis tasks. Specifically, this paper investigates text augmentation as a data enhancement strategy in which synthetic samples are generated from existing data to supplement the training set.

The dual challenges of data sparsity and the high cost of human annotation have spurred interest in data augmentation techniques. However, previous approaches have often relied on simple heuristics or rule-based systems, limiting their applicability and effectiveness. We posit that large language models like ChatGPT can serve as robust, general-purpose engines for generating high-quality synthetic data, thereby alleviating these limitations.

We conduct comprehensive experiments with transformer-based models, such as RoBERTa, across two different sentiment analysis datasets: PerSenT \cite{bastan2020author} and MultiEmo \cite{kocon2021multiemo}. The central questions we seek to address are twofold: First, can ChatGPT-augmented data boost the performance of smaller models to a level comparable with larger, more resource-intensive models? Second, do the benefits of using augmented data extend across different scales of models, from smaller to larger architectures?

By focusing on these questions, this paper aims to illuminate the potential of text augmentation via large language models, not only to improve model performance but also as a strategy for achieving computational efficiency and cost-effectiveness in AI applications.

Our list of contributions is as follows:
\begin{enumerate}
    \item We present an innovative approach that exploits ChatGPT's generative abilities to enhance the training data for sentiment analysis tasks. This strategy is poised to circumvent the challenges associated with collecting and annotating vast datasets.
    \item By leveraging ChatGPT for data augmentation, our methodology moves beyond the constraints of conventional heuristic and rule-based augmentation techniques, showcasing the potential for a more flexible and high-quality data enhancement.
    \item We conduct experiments with transformer models, particularly RoBERTa, on two sentiment analysis datasets (PerSenT and MultiEmo). These experiments probe into the efficiency and performance of models trained on ChatGPT-augmented data.
    \item The paper delves into pivotal questions concerning the scalability benefits of using augmented data and the feasibility of achieving high performance with smaller, less resource-intensive models.
    \item Our research emphasizes improving models' performance and underscores the importance of achieving computational thriftiness and cost-effectiveness in AI applications, especially in a world leaning towards on-device deployments.
\end{enumerate}

\section{Related Work}

%
\subsection{Text Augmentation Techniques}

The scarcity of data in machine learning tasks, particularly text classification, poses challenges for model training. A common remedy for this is text augmentation, where new data is artificially generated from the existing dataset to enrich the model's learning experience \cite{feng2021survey, shorten2021text, bayer2022survey}. Techniques for text augmentation are diverse and can be applied at various levels of granularity, as outlined below:

\begin{itemize}
\item \textbf{Character Level:} Methods at this level manipulate individual characters to generate new text \cite{belinkov2017synthetic, coulombe2018text, ebrahimi2017hotflip}. The advantage here is the subtlety of the changes, which can be nearly imperceptible yet effective in creating variations. However, the downside is that such minute alterations may sometimes lead to meaningless or garbled words, which may not aid in effective training.
\item \textbf{Word Level:} This involves techniques like synonym replacement, word swapping, and more to augment text at the lexical level \cite{wei2019eda, kolomiyets2011model, miller1990introduction, kocon2018context,kocon2018classifier,kobayashi2018contextual,kocon2019propagation,wierzba2021emotion,mao2023metapro}.  This method is beneficial as it maintains the overall sentence structure while introducing variability. The drawback, however, is that certain synonym replacements might change the context or nuance of the original sentence, which could be problematic for sensitive tasks.
\item \textbf{Sentence Level:} Sentence-level augmentation modifies or restructures existing sentences for new variants \cite{feng2019keep, min2020syntactic, kim2022alp, marivate2020improving,pogoda2022deep}. Their strength lies in generating more diverse sentence structures, thereby potentially enriching the dataset. However, there's a risk that excessive modification might divert from the original intent or sentiment of the source sentence, making the augmentation counterproductive.
\item \textbf{Document Level:} Entire documents are augmented to create data that maintains contextuality and coherence \cite{aiken2010efficacy, qiu2020easyaug, sun2020novel}. Their advantage is ensuring the augmented data still feels like a cohesive unit, which is particularly crucial for tasks that rely on understanding context over longer passages. On the flip side, these methods might be computationally more intensive and challenging to implement as maintaining contextuality across a full document can be complex.
\end{itemize}

In the present study, our primary focus is on document-level text augmentation.

\subsection{Advancements in Large Language Models (LLMs)}

Transformers have revolutionized the field of natural language processing \cite{vaswani2017attention}. Specifically, generative transformer models, known as large language models (LLMs), are becoming increasingly popular owing to their extensive parameter space \cite{radford2018improving}. Several variants exist:

\begin{itemize}
\item \textbf{GPT Series:} Including GPT-3 and GPT-4, these models are benchmarks in the field \cite{brown2020language}.
\item \textbf{FLAN and Bloom:} These models represent advancements in specific aspects like efficiency and capability \cite{longpre2023flan, scao2022bloom}.
\item \textbf{LLaMA Family:} Known for their robustness across multiple tasks \cite{touvron2302llama, touvron2023llama}.
\item \textbf{ChatGPT:} A specialized version of InstructGPT trained via reinforcement learning from human feedback (RLHF), known for conversational applications \cite{ouyang2022training, kocon2023chatgpt}.
\end{itemize}

This work will employ ChatGPT due to its adaptability and general-purpose capabilities.

\subsection{Data Generation and Augmentation via LLMs}

LLMs offer an intriguing avenue for generating and augmenting data. These models can produce data that closely aligns with human-generated content by feeding specific prompts. Some studies have utilized ChatGPT to generate entire synthetic datasets \cite{ubani2023zeroshotdataaug, yang2023neural}. Others have explored using LLMs for augmenting existing datasets \cite{sarker2023medical, dai2023chataug, whitehouse2023llm, maragheh2023llm}. These methods have generally led to improved performance in various machine-learning tasks.

\subsection{Deep Learning in Sentiment Analysis}
Sentiment analysis, the focus of this paper, is a well-studied problem aimed at classifying the sentiment conveyed in a text. Over the years, many methods and architectures have been proposed and implemented to decipher and categorize sentiments in textual data \cite{cambria2017affective,kocon2019multi,kocon2019multi2,kocon2019multilingual,kocon2019recognition,kanclerz2020cross,kocon2021ipm,kocon2021mapping,baran2022linguistic,kocon2021aspectemo,kocon2021multiemo,szolomicka2022multiaspectemo,korczynski2022compression,kocon2022neuro,kazienko2023human}. Ranging from early lexicon-based techniques to advanced pre-trained models, the field has seen vast developments.

\begin{itemize}
\item \textbf{Lexicon-based Techniques:} These techniques rely on sentiment dictionaries or lexicons, wherein words are pre-associated with sentiment scores. Tools such as SenticNet have further enhanced the lexicon-based approaches, combining commonsense reasoning with neuro-symbolic computations for a more holistic sentiment analysis \cite{cambria2022senticnet}.
\item \textbf{Recurrent Neural Networks:} Models such as BiLSTM/CNN and MCBiGRU are adept at capturing text-dependent sequential dependencies. Their ability to memorize previous information and context makes them suitable for sentiment analysis tasks \cite{chen2017improving}. A newer trend in sentiment analysis and comment toxicity detection, this method employs a combination of convolutional and recurrent architectures to process text from multiple channels, ensuring a more comprehensive understanding of the data \cite{kumar2021comment}.
\item \textbf{Transformer Models:} Techniques like BERT and RoBERTa have risen in prominence due to their unparalleled capability to capture contextual information. These architectures, especially when pre-trained, have set new standards in sentiment analysis benchmarks \cite{kocon2019multi, devlin2018bert, liu2019roberta, rybak2020klej,srivastava2023beyond,koptyra2023clarin}. An increasingly popular choice, these models, trained on vast amounts of text data, possess an inherent understanding of language semantics and structure. However, concerns regarding biases in these models have been highlighted in recent literature, emphasizing the need for careful and context-aware implementation \cite{kanclerz2021controversy,milkowski2021personal,kocon2021ipm,kocon2021learning,mao2022biases,milkowski2022multitask,kanclerz2022if,bielaniewicz2022deep,ngo2022studemo,ferdinan2023personalized,mieleszczenko2023capturing,kocon2023differential}.
\end{itemize}

Incorporating these diverse methods, this paper aims to make significant contributions to the field of sentiment analysis. A particular emphasis is placed on leveraging ChatGPT for data augmentation, bridging the gap between deep learning techniques and real-world sentiment analysis applications.

\section{Data Augmentation Methodology}

To augment our dataset, we utilized OpenAI's GPT-3.5 model via its API, employing default settings. Our method involves generating new data points based on the original dataset by using four distinct prompts. These prompts are designed to cover two different approaches to text augmentation: paraphrasing and generating entirely new text inspired by the original.

\subsection{Prompt-based Strategy}

Our approach adopts a two-tier structure for each augmentation method: \emph{paraphrasing} and \emph{inspirational generation}. However, it's essential to understand the differences and potential implications of each method:

\begin{itemize}
\item \textbf{Paraphrasing:} 
The primary goal of paraphrasing is to offer a varied representation of the same message or content. While it brings subtle variations to the text, it often doesn't exceed the inherent domain or context of the original content. This limitation can be both an advantage and a drawback. The advantage is that the paraphrased content remains tightly related to the original data, ensuring that the augmented data remains contextually relevant. However, the downside is that this might not sufficiently diversify the dataset, potentially limiting the enhanced generalization of the model.
\item \textbf{Inspirational Generation:} 
This strategy aims to craft content that is distinct from the original but maintains the same sentiment. The idea is to widen the scope of data to potentially traverse different domains or contexts, which might provide a more comprehensive augmentation. This method's primary advantage is that it might aid in creating a more domain-agnostic model, as the new texts could span diverse themes. However, there's a risk involved. As the generated content deviates from the original, there's a possibility that some generated examples might not align perfectly with the expected sentiment, potentially introducing noise into the dataset.
\end{itemize}

\subsection{Detailed Prompt Descriptions}
\label{sec:detailed_prompt}

For both augmentation strategies, we've delineated the process with specific prompts:

\begin{itemize}
\item For \textbf{Paraphrasing:} The first two prompts come under this umbrella. They are presented sequentially in a single conversation session with the model:
\begin{itemize}
    \item "Generate a paraphrase for the following text, preserving the sentiment of the following statement: \textit{text}"
    \item "Generate another paraphrase by changing more words also keeping the sentiment"
\end{itemize}
\item For \textbf{Inspirational Generation:} The latter two prompts are crafted to generate sentiment-consistent yet distinct content. They are initiated in individual conversation sessions with the model:
\begin{itemize}
    \item "Based on the given text, generate another text with a completely new theme, but be inspired by the original text and keep the sentiment of the old one in the new text. Original text: \textit{text}"
    \item "Based on the given text, generate another text with a completely new theme, but be inspired by the original text and keep the \textit{label} sentiment. Original text: \textit{text}"
\end{itemize}
\end{itemize}

\subsection{Implementation Details}

For every sample in each dataset, we applied all four prompts. It is important to note that:

\begin{itemize}
\item A single session was used for the first two prompts, wherein the second prompt was input only after the model responded to the first.
\item The third and fourth prompts were processed in separate sessions. The primary distinction between them is that the third prompt directs the model to infer the sentiment from the original text, whereas the fourth one includes the sentiment label explicitly in the prompt.
\end{itemize}

\subsection{Output Datasets}

We generated four separate augmented datasets by applying these prompts to each sample from the original training set. Each of these datasets maintains the class distribution of the original data, providing us with a richer resource for training our machine learning models.
\begin{taskbox}[myprompt]{Example augmentation}
\vspace*{-0.12cm}
\tcbsubtitle{Sample from the original dataset} 
With my children (10 years old and 2 years old) I actually only go to doctor. I think that she is a great specialist, she does not treat the patient "from above", she is nice, factual and well remembers and knows her little patients. She has an excellent attitude towards children and they like her very much. She does not prescribe immediately, sometimes without any sense, an antibiotic, only when it is necessary. A few times I happened to get to another pediatrician, but as a rule I was not satisfied. I'm glad that the doctor is working in my clinic. When he leaves, I'll have trouble choosing another pediatrician. I highly recommend it!
\tcbsubtitle{Paraphrased sample by the first prompt}
I only take my children (10 years old and 2 years old) to one doctor, and I believe she is an exceptional specialist. She doesn't treat her patients with condescension and has a warm and friendly demeanor. She demonstrates excellent knowledge of and familiarity with her young patients. It is evident that she genuinely cares about children and they have a strong liking for her. She only prescribes antibiotics when necessary and avoids unnecessary medication. Although I have occasionally visited other pediatricians, I have mostly been dissatisfied. I am grateful that this doctor is a part of my clinic, and I anticipate difficulty in choosing a new pediatrician when she eventually leaves. I highly recommend her!
\tcbsubtitle{Generated sample by the third prompt}  When it comes to planning family vacations, I usually rely on only one travel agency. Not just any agency, but one that I consider to be exceptional in providing personalized service and creating unforgettable experiences. They don't treat their clients as mere customers, but as cherished guests, ensuring every aspect of our trip is taken care of. From the moment we step foot into their office, their friendly and knowledgeable staff greet us with warmth and empathy. They have a remarkable understanding of our preferences and take note of even the smallest details that truly matter to us.  Unlike other agencies, they don't rush into booking the first available package. Instead, they carefully consider our desires and requirements before suggesting the perfect destination and itinerary. They genuinely care about our well-being and never recommend unnecessary add-ons or impulse purchases. I've occasionally tried other agencies, but I have always been left dissatisfied. It's the kind of satisfaction that lingers in your mind, making it difficult to choose another agency in the future.   I am truly grateful for the exceptional team at this agency. Their dedication, professionalism, and genuine love for what they do sets them apart. It's not just about business for them; it's about creating magical memories for families like mine. I can wholeheartedly recommend their services to anyone searching for a travel agency that goes above and beyond.
\tcbsubtitle{Sentiment label} 
Positive
\end{taskbox}
\section{Datasets}\label{sec:datasets}

For the sentiment analysis classification task, this paper employs two distinct datasets: PerSenT \cite{bastan2020author} and MultiEmo \cite{kocon2021multiemo}. Both datasets are well-suited for our research aims but differ in document length, language diversity, and class labels. A detailed statistical summary of each dataset is provided to understand their characteristics comprehensively.

\subsection{PerSenT}

The PerSenT dataset encompasses approximately 50,000 documents. On average, each document in this dataset comprises 377 words. The dataset is divided into three subsets: a training set with 3,355 documents, a validation set containing 578 documents, and a test set with 827 documents.

Texts in PerSenT are labeled according to one of three sentiment classes: \textit{Positive}, \textit{Negative}, and \textit{Neutral}. The distribution of these classes across the training, validation, and test sets can be found in Table~\ref{tab:dataset_distr}.

\subsection{MultiEmo}

MultiEmo is a more diverse dataset constructed from consumer reviews across four different domains: medicine, hotels, products, and universities. The entire corpus consists of documents in 11 languages, and each language subset contains roughly 8,000 documents. For the scope of this paper, we focus solely on the English documents combined from all domains.

In terms of document length, MultiEmo texts are generally shorter than those in PerSenT, averaging around 140 words. The dataset is divided into training, validation, and test subsets, containing 6,572, 823, and 820 texts respectively. Unlike PerSenT, MultiEmo includes a fourth sentiment class, \textit{Ambivalent}, along with \textit{Positive}, \textit{Negative}, and \textit{Neutral}. The class distribution is detailed in Table~\ref{tab:dataset_distr}.

\begin{table*}[h]
\caption{The datasets class distribution}
\begin{center}
\begin{tabular}{|c|c|c|c||c|c|c|c|}
\hline
\textbf{Dataset}&\multicolumn{3}{|c||}{\textbf{PerSenT}}&\multicolumn{4}{|c|}{\textbf{MultiEmo}}\\
\cline{2-8}
\textbf{split} & \textit{Positive} & \textit{Negative} & \textit{Neutral} & \textit{Positive} & \textit{Negative} & \textit{Neutral} & \textit{Ambivalent} \\
\hline
train & $52.4\%$ & $10.46\%$ & $37.14\%$ &$ 27.74\%$ & $37.57\% $& $14.77\%$ & $19.92$\%\\
\hline
valid & $52.6\% $& $10.03\%$ & $37.37\% $& $28.68\%$ & $36.94\%$ & $15.55\% $& $18.83\%$\\
\hline
test & $44.5\% $&$ 16.81\% $&$ 38.69\%$ & $27.68\% $& $41.34\%$ & $14.39\% $&$16.59\%$\\
\hline
\end{tabular}
\label{tab:dataset_distr}
\end{center}
\end{table*}

\section{Experiments}

\subsection{Setup}

The experiments were conducted using three transformer-based models: XLM-RoBERTa-comet-small \cite{wang2020minilmv2} (referred to as RoBERTa-small), XLM-RoBERTa-base \cite{conneau2019unsupervised} (referred to as RoBERTa-base), and XtremeDistil \cite{mukherjee2020xtremedistil, mukherjee2021xtremedistiltransformers} configured with six layers and a hidden size of 384. The parameter counts for these models are as follows:

\begin{itemize}
    \item RoBERTa-base: 279 million parameters
    \item RoBERTa-small: 107 million parameters
    \item XtremeDistil: 13 million parameters
\end{itemize}

\subsubsection{Data Augmentation Strategies}
We employed various original and augmented data combinations to form new training sets. Seven unique combinations were tested, each preserving the original class distribution.  For every augmentation strategy and specific prompt (Section~\ref{sec:detailed_prompt}), we generated synthetic data equivalent to the volume of the original training set.
\subsubsection{Baseline Experiment}
The baseline experiment used only the original training set without any form of data augmentation. All other experiments extended the original training set with augmented data generated using specific prompts, as detailed below:

\begin{itemize}
    \item \textit{Para}: Contains original training set and data augmented using the first prompt for paraphrasing.
    \item \textit{Para-Conv}: Contains original training set and data augmented using the second prompt for paraphrasing in a conversational context.
    \item \textit{Both Para}: Merges \textit{Para} and \textit{Para-Conv}.
    \item \textit{Insp}: Contains original training set and data augmented using the third prompt for generating inspired text.
    \item \textit{Insp-Lab}: Similar to \textit{Insp}, but uses the fourth prompt that includes sentiment labels.
    \item \textit{Both Insp}: Merges \textit{Insp} and \textit{Insp-Lab}.
    \item \textit{All}: Consolidates all augmented data with the original training set.
\end{itemize}

\subsubsection{Model Training Parameters}
All models were trained with a learning rate of \(10^{-5}\) -- a typical value often adopted when fine-tuning models, especially for transformer-based architectures \cite{liu2019roberta}. The batch size was set to 16, a decision primarily influenced by the memory limitations of our graphics card (RTX 3090). Tokenizers were specifically configured to truncate texts, capping them at a maximum of 512 tokens. This truncation length was selected because 512 tokens represent our model's maximum allowable input length.

\subsection{Experimental Results}

With the seven data combinations and three transformer models, 42 different experiments were conducted. Each experiment used the validation and test sets from the original data splits without modification. Each experimental run was performed five times with varying random seeds for reliability. The code is available on the GitHub repository\footnote{\url{https://github.com/CLARIN-PL/text-augumentation-with-chatgpt}}.

\subsection{Inference Time Evaluation}
To assess the inference efficiency of each model, we conducted a separate experiment measuring the time required for processing a sample batch of size 16 across 2,000 iterations. Average inference times for each model were calculated for both the PerSenT and MultiEmo datasets, as detailed in section \ref{sec:datasets}.

\subsection{Results}

\subsubsection{Evaluation Metrics}
Our evaluation of the experiments is based on three primary metrics: accuracy, F1 macro, and gain \cite{kocon2023chatgpt}. The gain metric quantifies the improvement of a given model relative to the baseline model for a specific metric, either accuracy or F1 score. It is calculated as shown in equation \ref{eq:gain}.

\begin{equation}\label{eq:gain}
    \text{Gain} = \frac{100\% \times \left(M - B\right)}{100\% - B}
\end{equation}

\subsubsection{PerSenT Dataset Results}
Results obtained from the PerSenT dataset are summarized in Table \ref{tab:average_results_persent}. These results show the F1 macro metric and accuracy averaged over 5 repetitions, alongside standard deviations. Generally, the augmented datasets yielded superior results compared to the baselines.
\begin{table}[htbp]
\caption{results persent}
\begin{center}
\begin{tabular}{|c|c|c|c|}
\hline
\textbf{Augmentation}&\multicolumn{3}{|c|}{\textbf{Transformer}} \\
\cline{2-4} 
\textbf{Type} & \textbf{RoBERTa-small} & \textbf{RoBERTa-base} & \textbf{XtremDistil} \\
\hline
& \multicolumn{3}{|c|}{\textbf{F1 macro}}\\
\hline
Baseline &  $ 36 \% \pm 2   \%$ &  $ 38 \% \pm 8  \%$ &  $ 41 \% \pm 3   \%$ \\
\hline
Para &  $ 38 \% \pm 3    \%$ &   $ 40 \% \pm 1  \%$ &  $ 41 \% \pm 2   \%$ \\
\hline
Para-Conv &  $ 39 \% \pm 1    \%$ &  $ 41 \% \pm 1   \%$ &  $ 43 \% \pm 2   \%$ \\
\hline
Both Para &   $ 40\% \pm 2    \%$ &   $ 37 \% \pm 1  \%$ &  $ 43 \% \pm 2   \%$ \\
\hline
Insp &  $ 37  \% \pm 4   \%$ &  $ 43 \% \pm 1   \%$ &  $ 43 \% \pm 3   \%$ \\
\hline
Insp-Lab &   $ 38  \% \pm 2   \%$ &  $ 41 \% \pm 2  \%$ &   $ 40 \% \pm 2   \%$ \\
\hline
Both Insp &   $ 37 \% \pm 1   \%$ &  $ 41 \% \pm 2  \%$ &  $ 42 \% \pm 3   \%$ \\
\hline
All &   $ 39 \% \pm 2    \%$ &  $ 39 \% \pm 1  \%$ &  $ 43 \% \pm 1   \%$ \\
\hline
& \multicolumn{3}{|c|}{\textbf{Accuracy}}\\
\hline
Baseline  &  $ 38 \% \pm 2   \%$ &  $ 39 \% \pm 8  \%$ &  $ 43 \% \pm 5   \%$ \\
\hline
Para   &  $ 39 \% \pm 4 \%  $   &   $ 46 \% \pm 1  \%$ &  $ 43 \% \pm 3   \%$ \\
\hline
Para-Conv &  $ 39 \% \pm 1    \%$ &  $ 44 \% \pm 1   \%$ &  $ 44 \% \pm 2   \%$ \\
\hline
Both Para &  $ 41  \% \pm 3   \%$ &   $ 45 \% \pm 3  \%$ &  $ 45 \% \pm 3   \%$ \\
\hline
Insp &  $ 37  \% \pm 5   \%$ &  $ 46 \% \pm 1   \%$ &  $ 44 \% \pm 3   \%$ \\
\hline
Insp-Lab &   $ 41  \% \pm 2   \%$ &  $ 46 \% \pm 4  \%$ &   $ 42 \% \pm 2   \%$ \\
\hline
Both Insp &   $ 40 \% \pm 1   \%$ &  $ 46 \% \pm 2  \%$ &  $ 44 \% \pm 3   \%$ \\
\hline
All &   $ 44 \% \pm 2    \%$ &  $ 46 \% \pm 2  \%$ &  $ 46 \% \pm 1   \%$ \\
\hline
\end{tabular}
\label{tab:average_results_persent}
\end{center}
\end{table}

\subsubsection{MultiEmo Dataset Results}
Metrics from the MultiEmo dataset are presented in Table \ref{tab:average_results_multiemo}. Though the dataset is less complex than PerSenT, we still observed improvements when using augmented data.
\begin{table}[htbp]
\caption{results multiemo}
\begin{center}
\begin{tabular}{|c|c|c|c|}
\hline
\textbf{Augmentation}&\multicolumn{3}{|c|}{\textbf{Transformer}} \\
\cline{2-4} 
\textbf{Type} & \textbf{RoBERTa-small} & \textbf{RoBERTa-base} & \textbf{XtremDistil} \\
\hline
& \multicolumn{3}{|c|}{\textbf{F1 macro}}\\
\hline
Baseline &  $ 78 \% \pm 2   \%$ &  $ 87 \% \pm 1  \%$ &  $ 85 \% \pm 1   \%$ \\
\hline
Para &  $ 83 \% \pm   0  \%$ &  $ 88 \% \pm 1  \%$ & $ 86 \% \pm 1   \%$ \\
\hline
Para-Conv & $ 82 \% \pm 1    \%$ & $ 87 \% \pm 1   \%$ & $ 87 \% \pm 1   \%$ \\
\hline
Both Para & $ 83  \% \pm 1   \%$ &  $ 87 \% \pm 1  \%$ & $ 86 \% \pm  0  \%$ \\
\hline
Insp &  $ 82  \% \pm 1   \%$ &  $ 86 \% \pm 1   \%$ &  $ 85 \% \pm 0   \%$ \\
\hline
Insp-Lab &   $ 82  \% \pm   0 \%$ &  $ 87 \% \pm  0 \%$ &   $ 85 \% \pm 1   \%$ \\
\hline
Both Insp &   $ 84 \% \pm 1   \%$ &  $ 86 \% \pm 1  \%$ &  $ 85 \% \pm 0   \%$ \\
\hline
All &   $ 85 \% \pm 1    \%$ &  $ 86 \% \pm 1  \%$ &  $ 87 \% \pm 1   \%$ \\
\hline
& \multicolumn{3}{|c|}{\textbf{Accuracy}}\\
\hline
Baseline &  $ 80 \% \pm   1   \%$ &  $ 88 \% \pm  1  \%$ &  $ 86 \% \pm  1   \%$ \\
\hline
Para &  $ 84 \% \pm   1    \%$ &   $ 89 \% \pm  1  \%$ &  $ 87 \% \pm  1   \%$ \\
\hline
Para-Conv &  $ 83 \% \pm  1    \%$ &  $ 88 \% \pm  1   \%$ &  $ 87 \% \pm  1   \%$ \\
\hline
Both Para &  $ 84  \% \pm   1   \%$ &   $ 88 \% \pm  1  \%$ &  $ 80 \% \pm   0 \%$ \\
\hline
Insp &  $ 84\% \pm   1   \%$ &  $ 87 \% \pm  1   \%$ &  $ 86 \% \pm   0 \%$ \\
\hline
Insp-Lab &   $ 84 \% \pm   1   \%$ &  $ 88 \% \pm  1  \%$ &   $ 86 \% \pm  1   \%$ \\
\hline
Both Insp &   $ 84 \% \pm   1   \%$ &  $ 87 \% \pm  1  \%$ &  $ 86 \% \pm   0 \%$ \\
\hline
All &   $ 85 \% \pm  1    \%$ &  $ 88 \% \pm  1  \%$ &  $ 88 \% \pm  1   \%$ \\
\hline
\end{tabular}
\label{tab:average_results_multiemo}
\end{center}
\end{table}

\subsubsection{Comparative Gain Analysis}
\begin{figure}
\centerline{\includegraphics[width=\linewidth]{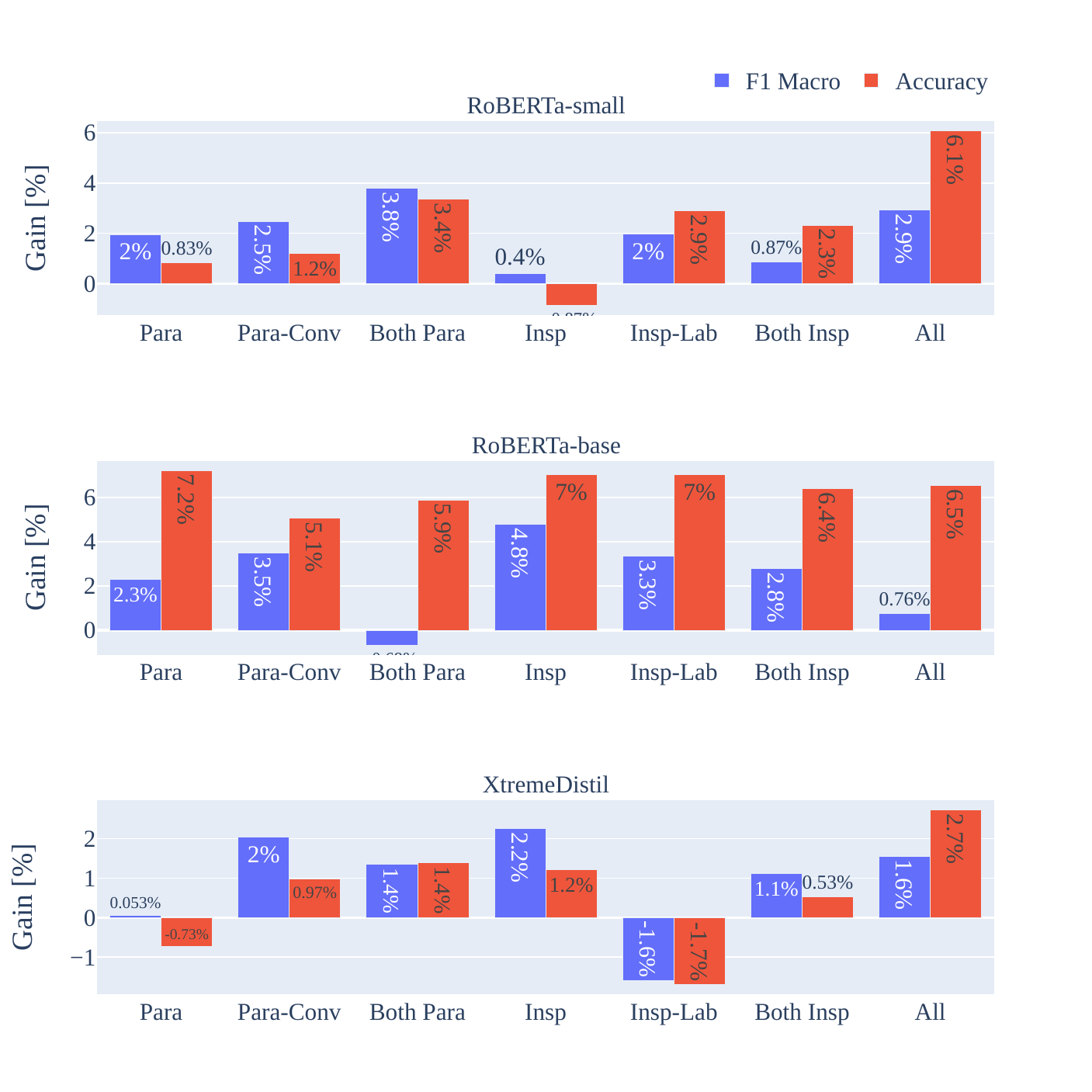}}
\caption{Gain metric per augmentation type on PerSenT dataset in accuracy and F1 macro metrics}
\label{fig:gain_persent}
\end{figure}

\begin{figure}
\centerline{\includegraphics[width=\linewidth]{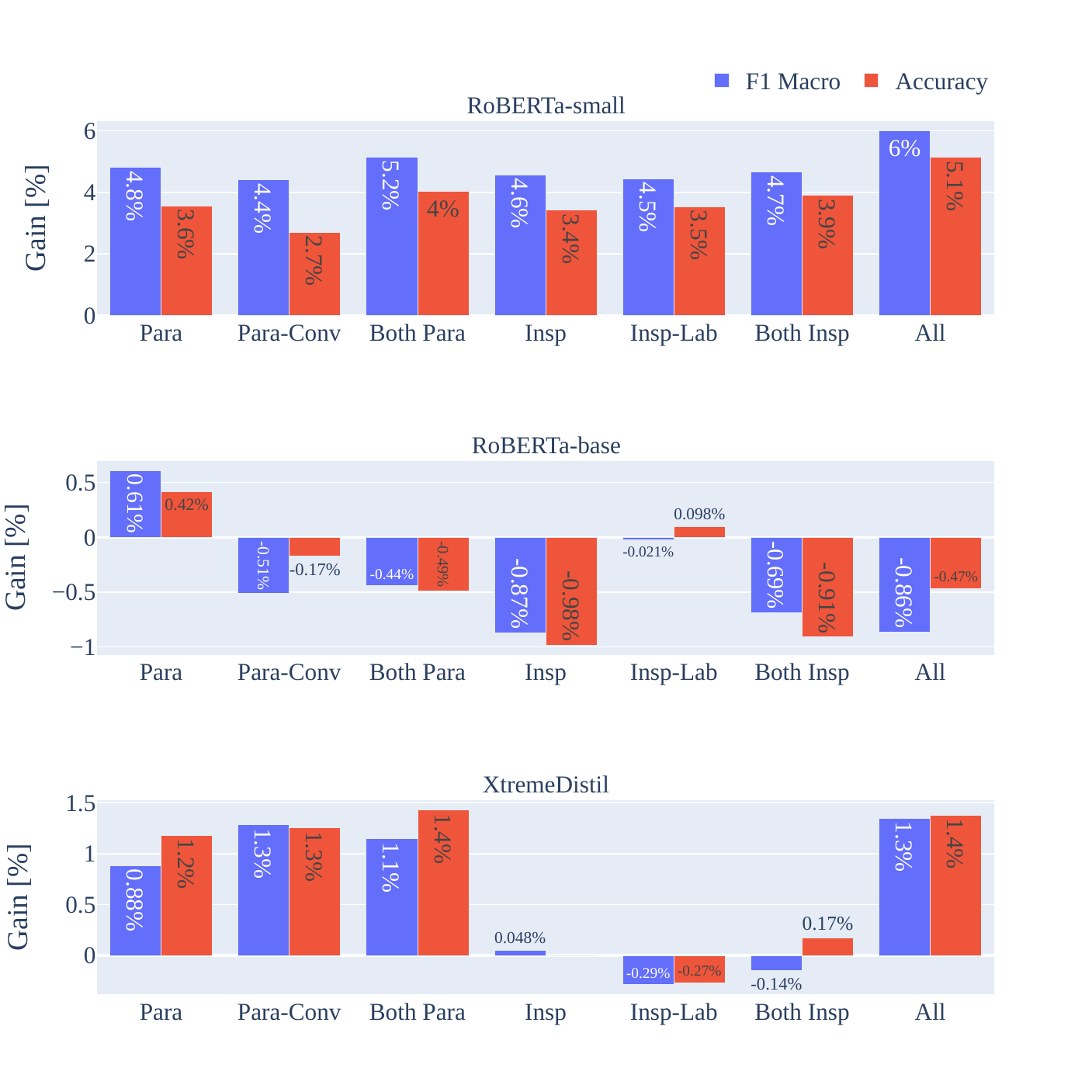}}
\caption{Gain metric per augmentation type on MultiEmo dataset in accuracy and F1 macro metrics}
\label{fig:gain_multiemo}
\end{figure}

The gain metric, calculated per equation \ref{eq:gain}, was computed for all experiments. Figure \ref{fig:gain_persent} and Figure \ref{fig:gain_multiemo} illustrate the gains for PerSenT and MultiEmo datasets, respectively.

In the less complex MultiEmo dataset, larger models often exhibited negative gains, while smaller models displayed positive gains. However, in the more complex PerSenT dataset, RoBERTa-base showed substantial improvements over the baseline.

\subsubsection{Class-Level Gain Analysis}
\begin{figure}
\centerline{\includegraphics[width=\linewidth]{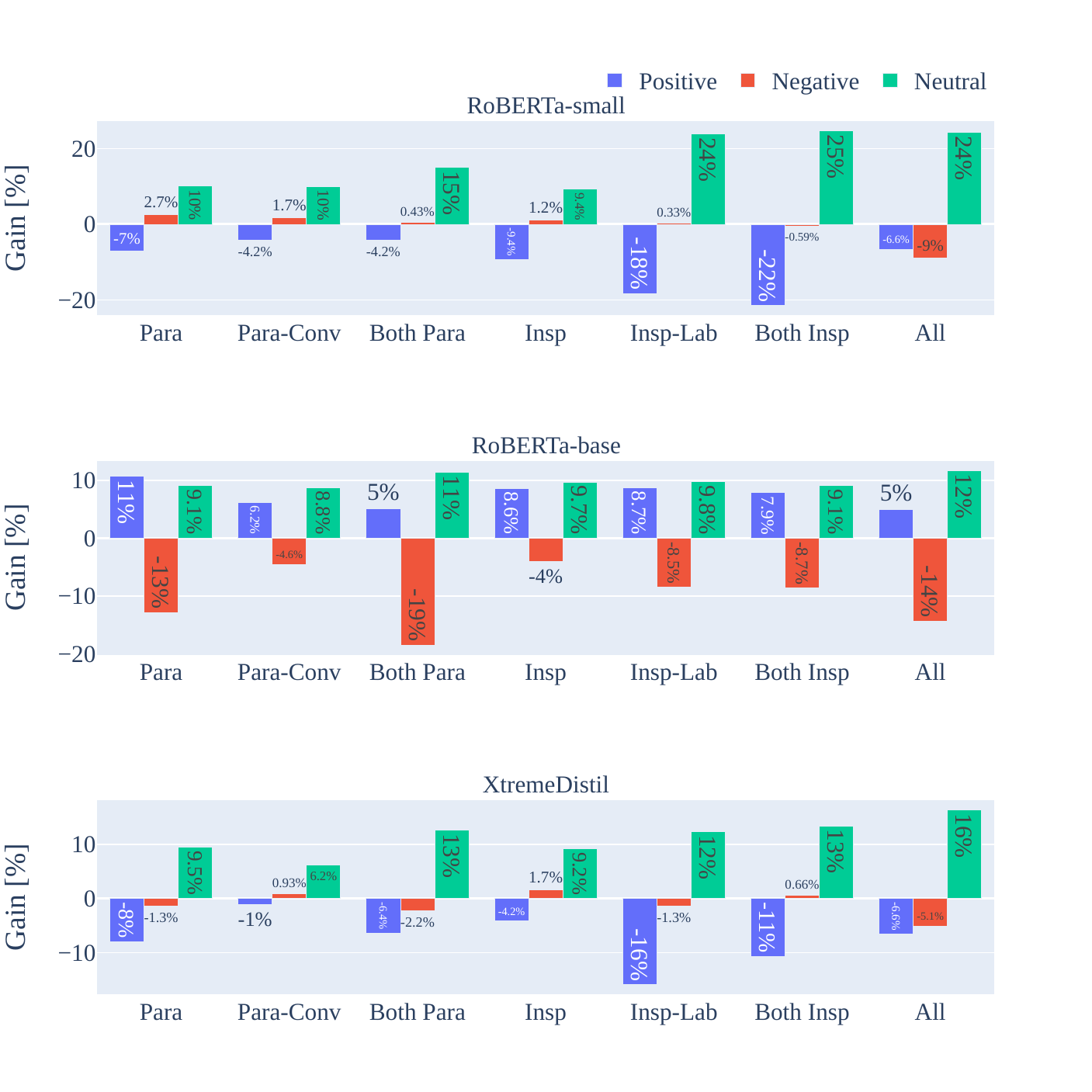}}
\caption{Gain metric per class and augmentation type on PerSenT dataset in F1 macro}
\label{fig:gain_persent_f1}
\end{figure}

\begin{figure}
\centerline{\includegraphics[width=\linewidth]{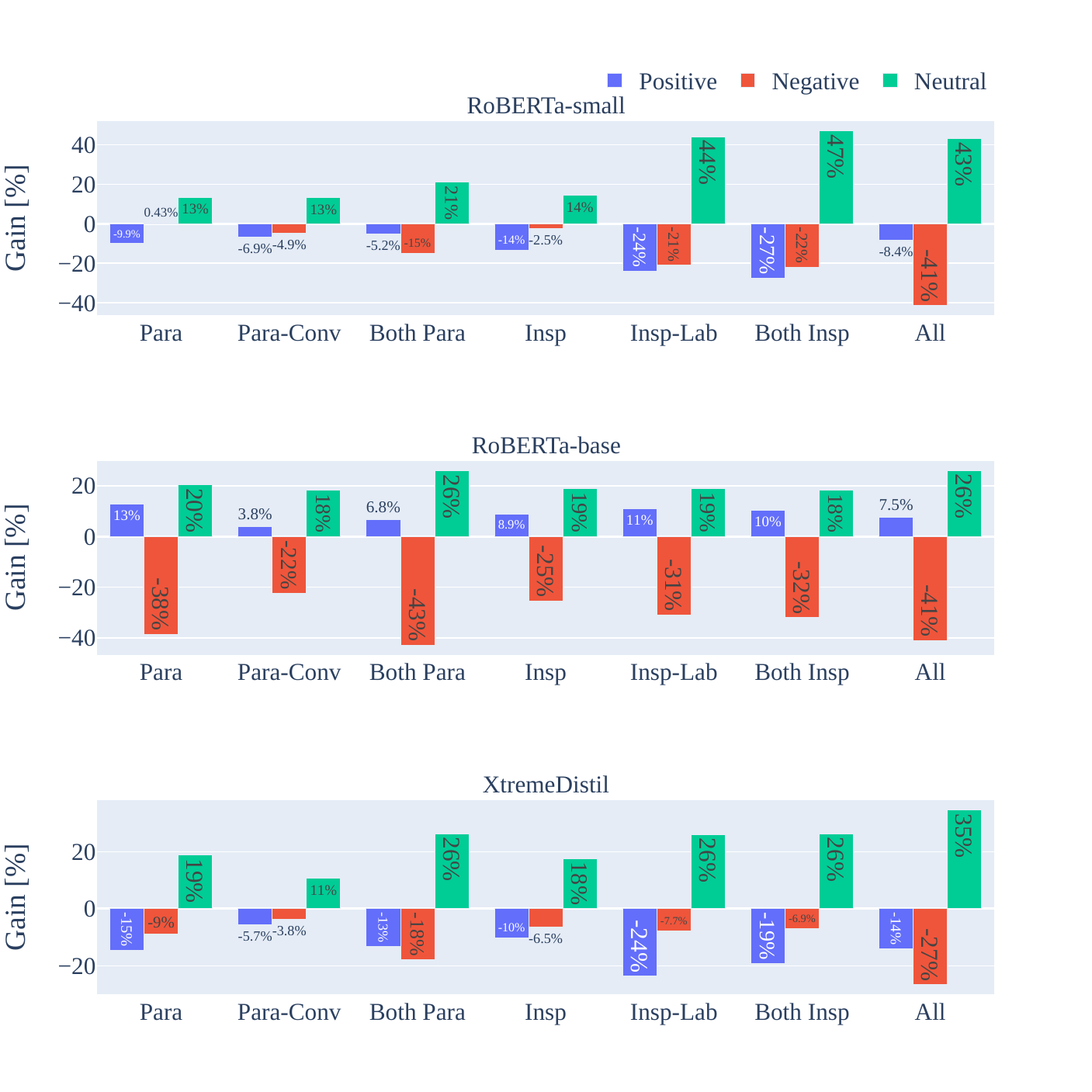}}
\caption{Gain metric per class and augmentation type on PerSenT dataset in accuracy}
\label{fig:gain_persent_acc}
\end{figure}

\begin{figure}
\centerline{\includegraphics[width=\linewidth]{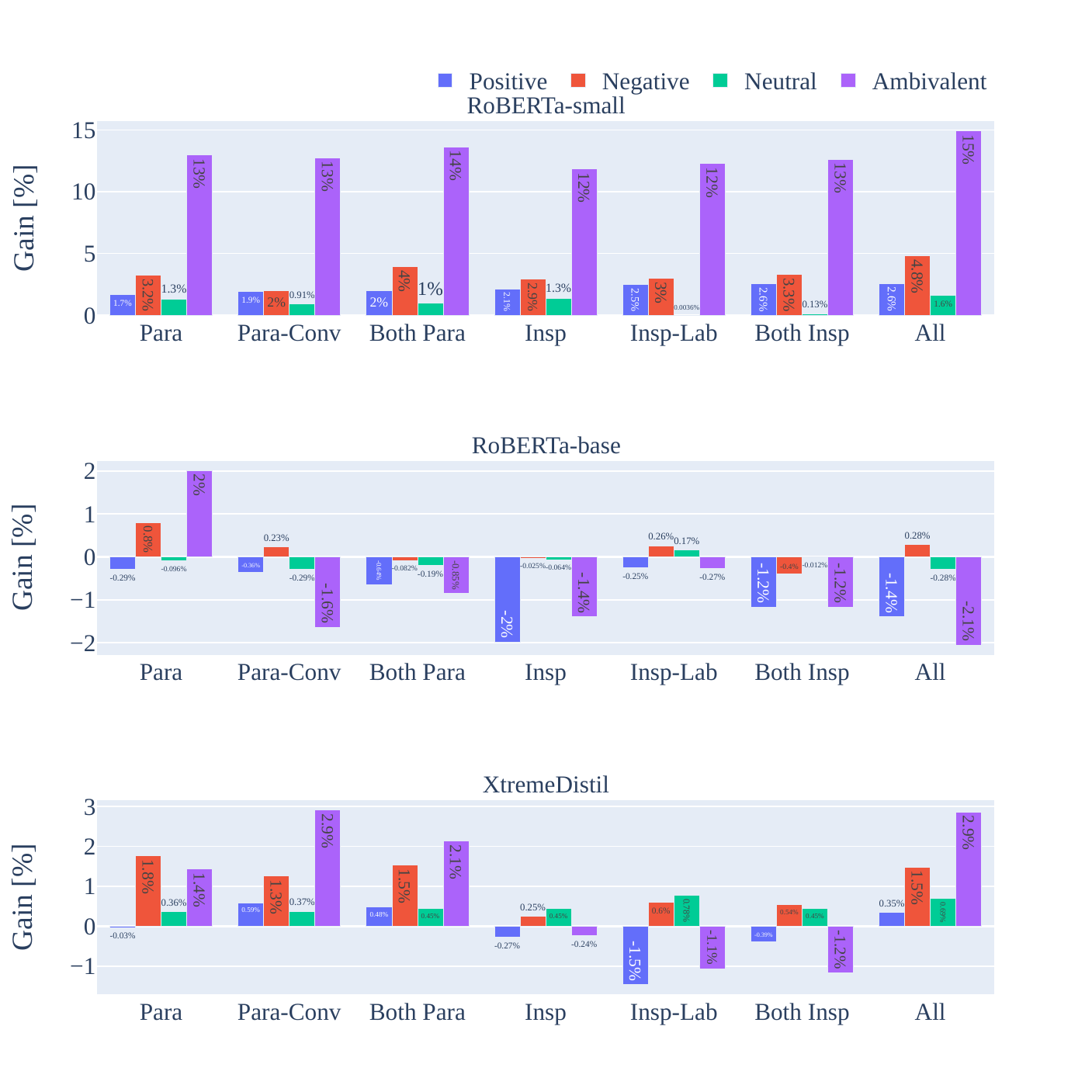}}
\caption{Gain metric per class and augmentation type on MultiEmo dataset in F1 macro}
\label{fig:gain_multiemo_f1}
\end{figure}

\begin{figure}
\centerline{\includegraphics[width=\linewidth]{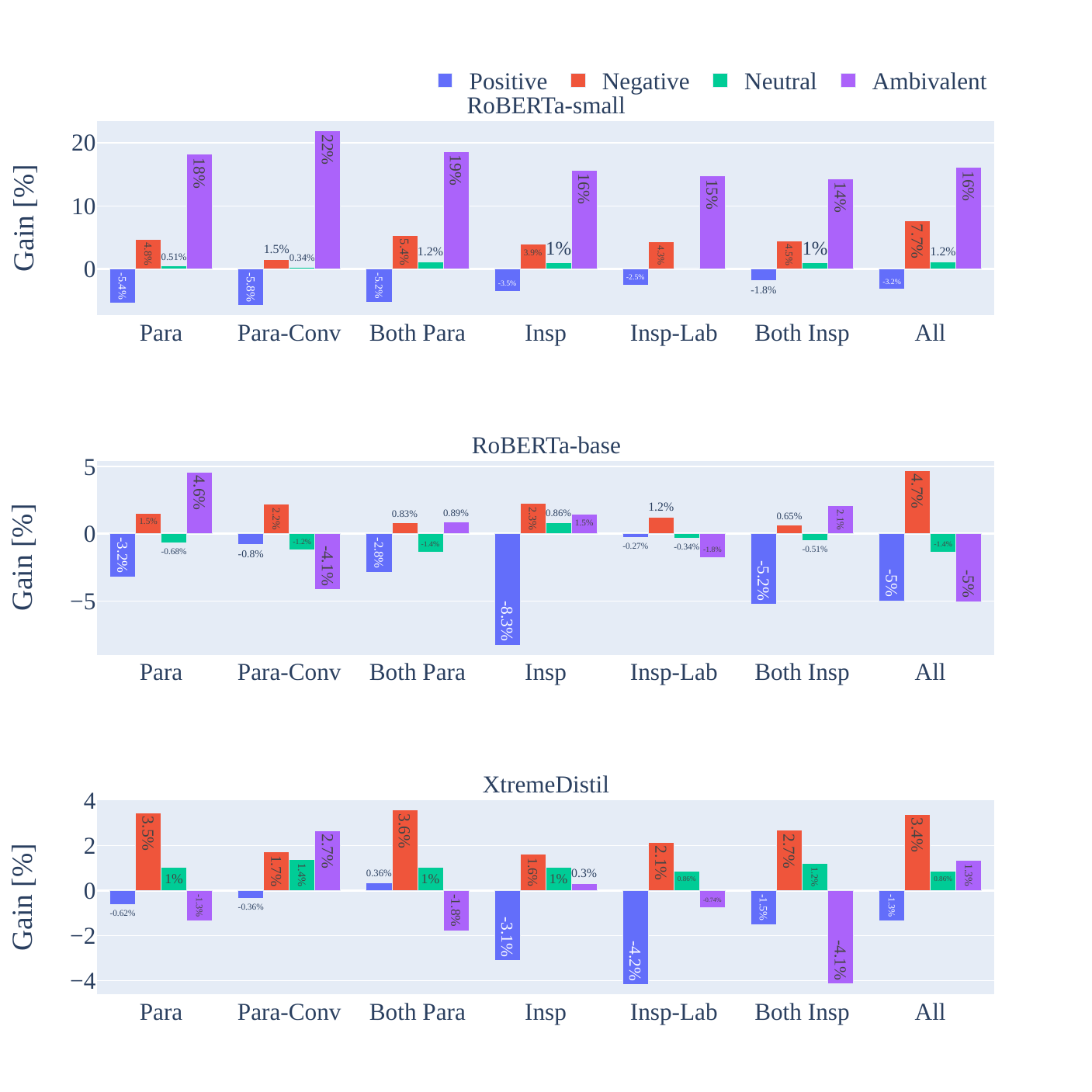}}
\caption{Gain metric per class and augmentation type on MultiEmo dataset in accuracy}
\label{fig:gain_multiemo_acc}
\end{figure}

We further delved into the gain metric at the class level. Figures \ref{fig:gain_persent_f1}, \ref{fig:gain_persent_acc}, \ref{fig:gain_multiemo_f1}, and \ref{fig:gain_multiemo_acc} display these metrics for both datasets. Interesting observations were made, such as significant improvements in the \textit{Ambivalent} class with RoBERTa-small on the MultiEmo dataset, with little to no trade-offs in other classes.

\subsubsection{Comparison with Baseline}
\begin{figure}
\centerline{\includegraphics[width=\linewidth]{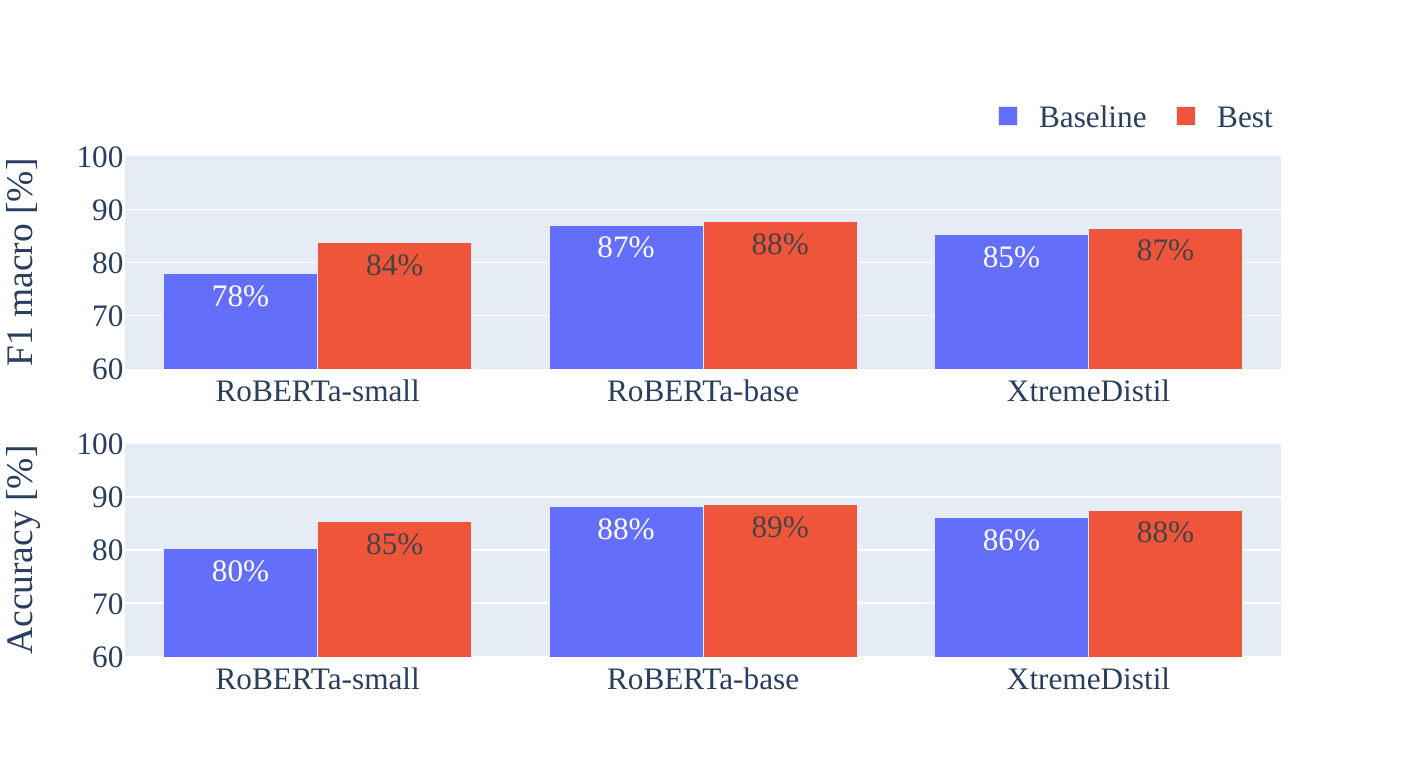}}
\caption{Baseline vs. best model on Multiemo dataset}
\label{fig:base_best_multiemo}
\end{figure}

\begin{figure}
\centerline{\includegraphics[width=\linewidth]{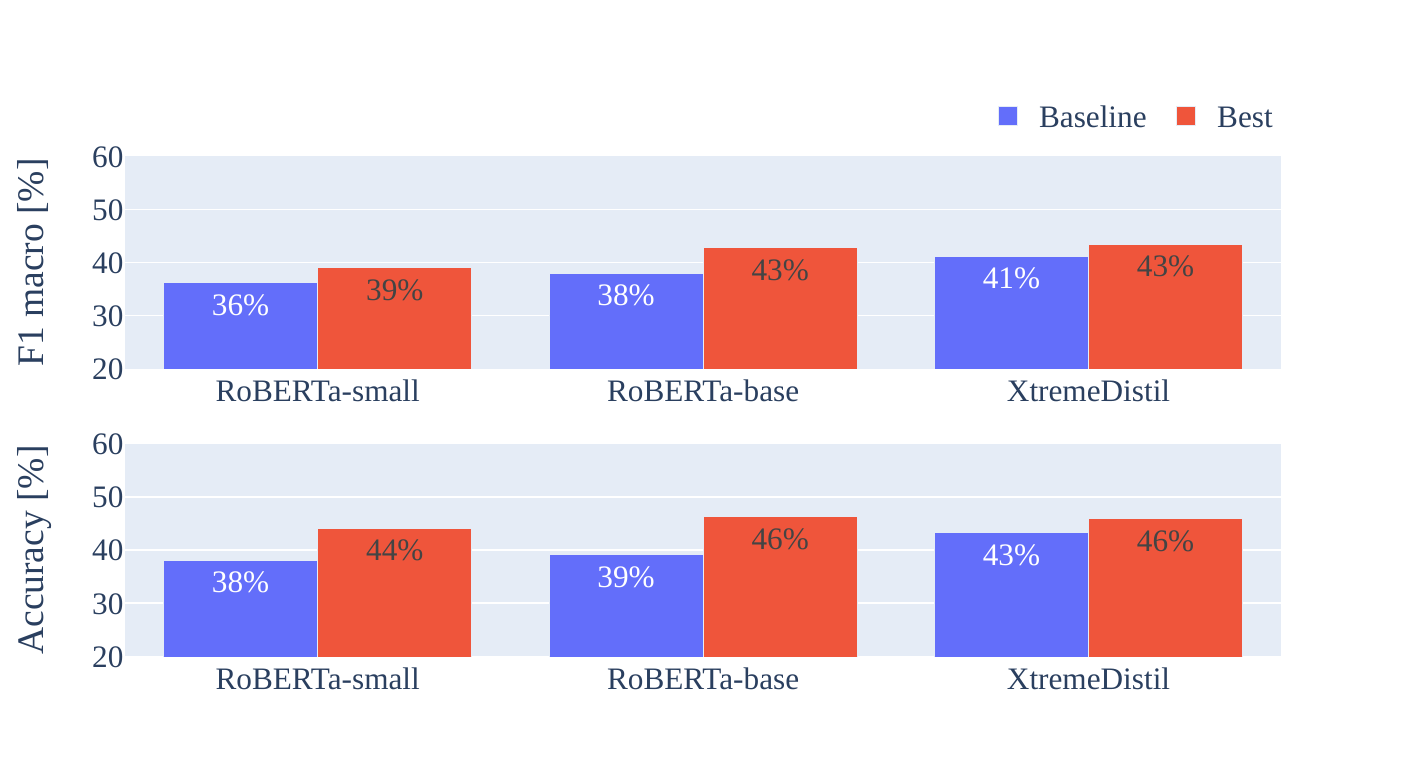}}
\caption{Baseline vs. best model on PerSenT dataset}
\label{fig:base_best_persent}
\end{figure}

Figures \ref{fig:base_best_multiemo} and \ref{fig:base_best_persent} contrast the best-performing models trained on augmented datasets against their corresponding baselines. In almost all cases, augmented training data improved model performance, bringing smaller models close to or surpassing the performance of larger models trained on original data.

\subsubsection{Inference Time Analysis}
\begin{figure}
\centerline{\includegraphics[width=\linewidth]{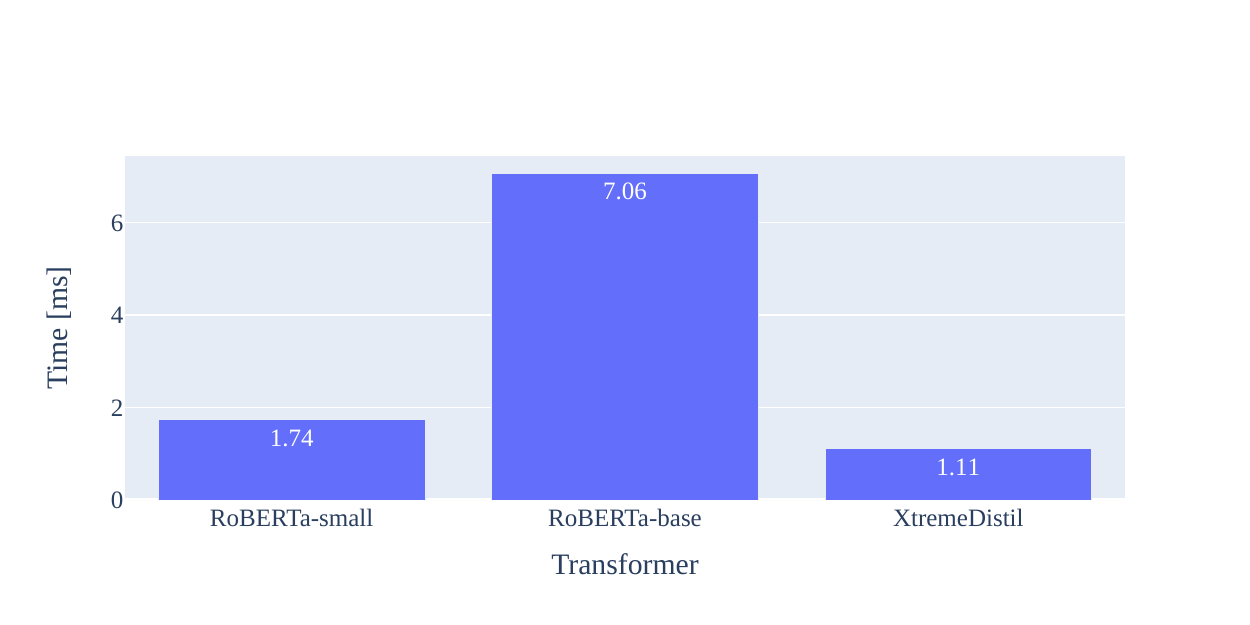}}
\caption{Inference time per sample}
\label{fig:time}
\end{figure}

Inference time results are presented in Figure \ref{fig:time}. As observed, RoBERTa-base had the longest inference times—almost seven times slower than XtremeDistil. It is worth noting that the time metrics were consistent across datasets, even though the MultiEmo texts were considerably shorter on average, thus confirming text length as a variable influencing inference time.

\subsection{Conclusions}

\subsubsection{Data Augmentation Efficacy}
The results affirm that data augmentation, specifically using large language models (LLMs), enhances the performance of transformer models in the tested tasks \cite{wang2020minilmv2,conneau2019unsupervised,mukherjee2020xtremedistil}. However, the effectiveness of augmentation varied depending on the type and the dataset in question.

\subsubsection{Optimization of Resources}
Our experiments highlight an important trade-off between computational resources and model performance. Training larger models, particularly with augmented data, can be resource-intensive and costly. Yet, our results suggest that smaller models trained on augmented datasets can achieve competitive or even superior performance while substantially reducing computational requirements \cite{wang2020minilmv2,mukherjee2020xtremedistil}.

\subsubsection{Performance Efficiency}
Smaller models benefited from speed-ups of nearly 7x during inference and saved up to 20 times the memory in terms of the number of parameters without compromising on performance quality. This suggests that adopting smaller, more efficient models is a viable strategy, especially when coupled with augmented training data \cite{wang2020minilmv2,mukherjee2020xtremedistil}.

\subsubsection{Class-Level Performance}
An intriguing observation was the differential impact of data augmentation on class-level performance. Larger models, particularly when trained on augmented datasets, demonstrated more significant performance degradation in specific classes without commensurate gains in other classes. This finding strengthens the argument for the utility of smaller models trained on augmented data, as they tend to offer a more balanced improvement across classes \cite{kocon2023chatgpt}.

\subsubsection{Future Directions}
While our work lays a foundational understanding of the benefits of data augmentation and model selection, ample room remains for exploration. Future work could investigate the effects of more nuanced forms of augmentation or delve deeper into the relationship between augmentation and the class distribution in datasets.

\subsubsection{Overall Summary}
In summary, our research demonstrates that strategic data augmentation can improve model performance while mitigating the challenges associated with resource-intensive larger models. This study paves the way for more efficient and cost-effective machine learning models, highlighting the merits of smaller, augmented models over their larger counterparts.

\section{Discussion and Future Work}

As the realm of artificial intelligence continues to expand, the imperative for deploying models on diverse platforms, including resource-constrained devices such as mobile phones, becomes increasingly salient. This paper has demonstrated that it is possible to maintain high-performance levels with significantly smaller models, thereby enabling their use on devices with limited memory without compromising the quality of results. Furthermore, the compact nature of these models enables on-device inference, thereby augmenting user privacy and security, which are critical concerns in today's digital landscape.

In the present study, we have concentrated on data augmentation utilizing GPT-3.5 as the large language model (LLM) of choice. Moving forward, several avenues for exploration and enhancement present themselves. One such avenue would be to investigate other LLMs for their potential in augmenting data. Alternative LLMs could offer superior capabilities in sentiment prediction, thereby enhancing the quality of the augmented dataset. Furthermore, some models might offer free API access without sacrificing quality, which could reduce costs substantially. The speed of the chosen model could also be a significant factor, as faster models would allow for quicker data collection, leading to potential cost savings.

Another area ripe for future inquiry is the optimization of prompt engineering for the selected LLM. Crafting more efficient and targeted prompts could yield augmented data of even higher quality. Beyond mere paraphrasing or simple data extensions, the use of carefully designed prompts could produce a broader range of text augmentations. These might include more nuanced reinterpretations of the original data, thus expanding the scope and diversity of the generated dataset. Additionally, more efficient prompt engineering could produce multiple, high-quality augmentations within a single model response, thereby enriching the dataset without a corresponding increase in processing time.

Lastly, we propose future experiments that explore the integration of original training sets with their augmented counterparts, specifically for class balancing. By merging the two types of datasets, it might be possible to eliminate the issues of class imbalance, thereby yielding more reliable and generalizable results at the class level.

In summary, the work presented herein opens up several promising avenues for future research, ranging from LLM selection and prompt engineering to dataset integration for class balancing. These endeavors could not only improve the efficiency and cost-effectiveness of utilizing LLMs for data augmentation but also significantly advance the field's understanding of optimizing model performance on resource-constrained platforms.

\section*{Acknowledgements}
This work was financed by 
(1) contribution to the European Research Infrastructure "CLARIN ERIC - European Research Infrastructure Consortium: Common Language Resources and Technology Infrastructure", 2022-23 (CLARIN Q);
(2) the European Regional Development Fund, as a part of the 2014-2020 Smart Growth Operational Programme, projects no. POIR.04.02.00-00C002/19, POIR.01.01.01-00-0923/20, POIR.01.01.01-00-0615/21, and POIR.01.01.01-00-0288/22; 
(3) the statutory funds of the Department of Artificial Intelligence, Wroclaw University of Science and Technology;
(4) the Polish Ministry of Education and Science within the programme “International Projects Co-Funded”;
(5) the European Union under the Horizon Europe, grant no. 101086321 (OMINO). However, the views and opinions expressed are those of the author(s) only and do not necessarily reflect those of the European Union or the European Research Executive Agency. Neither the European Union nor European Research Executive Agency can be held responsible for them.

\bibliography{main}


\end{document}